\PassOptionsToPackage{usenames,dvipsnames,table}{xcolor}
\documentclass[11pt,a4paper]{article}
\usepackage[hyperref]{emnlp2020}
\usepackage{times}
\usepackage{latexsym}

\usepackage{microtype}

\aclfinalcopy 

\usepackage{times}
\usepackage{latexsym}
\usepackage{url}
\usepackage{amssymb}
\usepackage{array}
\usepackage{amsmath}
\usepackage{algorithm}
\usepackage[noend]{algpseudocode}
\usepackage{graphicx}
\usepackage{url}
\usepackage{times}
\usepackage{helvet}
\usepackage{courier}
\usepackage{balance}  
\usepackage{mdwlist}
\usepackage{graphics}
\usepackage{color}
\usepackage{rotating}
\usepackage{booktabs}
\usepackage{epsfig}
\usepackage{alltt}
\usepackage{tabularx}
\usepackage{cellspace}
\usepackage{array}
\usepackage{moreverb}
\usepackage{fancyvrb}
\usepackage{enumerate}
\usepackage{colortbl}           
\usepackage{xspace}
\usepackage{relsize}
\usepackage{amsfonts}
\usepackage{txfonts}
\usepackage[labelfont=bf]{caption}
\usepackage{dsfont}
\usepackage{rotating}
\usepackage{latexsym}
\usepackage{multirow}
\usepackage{booktabs}
\usepackage{float}
\usepackage{array,etoolbox}
\usepackage{enumitem}
\usepackage{subfig}
\usepackage{colortbl}
\usepackage{arydshln}
\usepackage{setspace}
\usepackage[page]{appendix}
\usepackage{multirow}
\usepackage{seqsplit}
\usepackage{mathptmx}
\usepackage{anyfontsize}
\usepackage{t1enc}
\usepackage{adjustbox}
\usepackage{chngpage}
\usepackage{amsmath}
\usepackage{bbm}
\usepackage{makecell}
\usepackage{ragged2e}
 \usepackage{afterpage}
 \usepackage{capt-of}
\usepackage{amsbsy}

\usepackage{ulem}
\usepackage{longtable} 
\usepackage{supertabular}

\usepackage{color}
\usepackage[dvipsnames]{xcolor}

\usepackage{etoolbox}
\AtBeginEnvironment{quote}{\singlespacing\small}

\newtoggle{draft}
\toggletrue{draft}
\iftoggle{draft}{
\newcommand{\marti}[1]{\textcolor{Blue}{[#1 \textsc{--marti}]}}
\newcommand{\dk}[1]{\textcolor{Maroon}{[#1 \textsc{--dk}]}}
\newcommand{\andrew}[1]{\textcolor{Green}{[#1 \textsc{--andrew}]}}
\newcommand{\kyle}[1]{\textcolor{Orange}{[#1 \textsc{--kyle}]}}

\newcommand{\risham}[1]{\textcolor{Brown}{[#1 \textsc{--risham}]}}
}{
\newcommand{\marti}[1]{}
\newcommand{\dk}[1]{}
\newcommand{\andrew}[1]{}
\newcommand{\kyle}[1]{}
\newcommand{\risham}[1]{}
}

\newcommand{\com}[1]{}

\newcommand{\method}{\textsc{HEDDE}x\xspace}

\let\appendixpagenameorig\appendixpagename
\renewcommand{\appendixpagename}{\sffamily\appendixpagenameorig}

\usepackage{cleveref}
\usepackage{commath}
\usepackage{fdsymbol}
\newcommand\tab[1][5mm]{\hspace*{#1}}
\crefformat{section}{\S#2#1#3} 
\crefformat{subsection}{\S#2#1#3}
\crefformat{subsubsection}{\S#2#1#3}

\newcommand{\heart}{\ensuremath\heartsuit}

\title{Document-Level Definition Detection in Scholarly Documents: \\Existing Models, Error Analyses, and Future Directions}

\makeatletter
\patchcmd{\NAT@test}{\else \NAT@nm}{\else \NAT@nmfmt{\NAT@nm}}{}{}
\DeclareRobustCommand\citepos
  {\begingroup
   \let\NAT@nmfmt\NAT@posfmt
   \NAT@swafalse\let\NAT@ctype\z@\NAT@partrue
   \@ifstar{\NAT@fulltrue\NAT@citetp}{\NAT@fullfalse\NAT@citetp}}
\let\NAT@orig@nmfmt\NAT@nmfmt
\def\NAT@posfmt#1{\NAT@orig@nmfmt{#1's}}
\makeatother

\author{Dongyeop Kang$^\heart$ \tab Andrew Head$^\heart$ \tab Risham Sidhu$^\heart$ \\
\textbf{Kyle Lo}$^\diamondsuit$ \tab \textbf{Daniel S. Weld}$^{\diamondsuit\circ}$ \tab \textbf{Marti A. Hearst}$^\heart$ \\
  $^\heart$University of California, Berkeley,
  $^\diamondsuit$Allen Institute for AI, 
  $^\circ$University of Washington\\
  \texttt{\{dongyeopk,andrewhead,rishamsidhu,hearst\}}\texttt{@berkeley.edu} \\
  \texttt{\{kylel,danw\}}\texttt{@allenai.org} 
 }

\date{}

\begin{document}
\maketitle
\begin{abstract}

The task of definition detection is important for scholarly papers, because papers often make use of  technical terminology that may be unfamiliar to readers. Despite prior work on definition detection, current approaches are far from being accurate enough to use in real-world applications.  

In this paper, we first perform in-depth error analysis of the current best performing definition detection system and discover major  causes of  errors. Based on this analysis, we develop a new definition detection system, \method, that utilizes syntactic features, transformer encoders, and heuristic filters, and evaluate it on a standard sentence-level benchmark.  Because current benchmarks evaluate randomly sampled sentences, we propose an alternative evaluation that assesses every sentence within a document.  This  allows for evaluating  recall in addition to precision.

\method outperforms the leading system on both the sentence-level and the document-level tasks, by 12.7 F1 points and 14.4 F1 points, respectively.
We  note that performance on the high-recall document-level task is much lower than in the standard evaluation approach, due to the necessity of incorporation of document structure as features. 
We discuss remaining challenges in document-level definition detection, ideas for improvements, and potential issues for the development of reading aid applications.

\end{abstract}

\section{Introduction}

Automatic definition detection is an important task in natural language processing (NLP). Definitions can be used for a variety of downstream tasks, such as ontology matching and construction \cite{bovi2015large}, paraphrasing \cite{hashimoto2011extracting}, and word sense disambiguation \cite{banerjee2002adapted,huang-etal-2019-glossbert}.  Prior work in automated definition detection has addressed the domain of scholarly articles  \cite{reiplinger-etal-2012-extracting,jin-etal-2013-mining,espinosa-anke-schockaert-2018-syntactically,vanetik-etal-2020-automated,Veyseh2020AJM}. Definition detection is especially important for scholarly papers because they often use  unfamiliar technical terms that readers must understand to properly comprehend the article.

\begin{table}[t!]
\footnotesize
\centering
\renewcommand{\arraystretch}{0.8}
\begin{tabular}{@{}l@{}}
\toprule
\textbf{Example} 
\\ \midrule \\
\begin{tabular}[c]{@{}l@{}} \uline{$s^{task}$} are \dashuline{softmax-normalized weights}  and the scalar [...] 
 \end{tabular}
 
 \\ \\ \hdashline[0.4pt/2pt] \\

\begin{tabular}[c]{@{}l@{}} \uline{Textual entailment} is \dashuline{the task of determining whether} \\ \dashuline{ a ``hypothesis'' is true, given a ``premise''.} \end{tabular} 

\\ \\ \hdashline[0.4pt/2pt] \\

\begin{tabular}[c]{@{}l@{}} A \uline{biLM} \dashuline{combines both a forward and backward LM}  \end{tabular} 

\\ \\ \hdashline[0.4pt/2pt] \\

\begin{tabular}[c]{@{}l@{}} [...] a fine grained \dashuline{word sense disambiguation} (\uline{WSD}) \\  task and a POS tagging task. \end{tabular} 

\\ \\

\bottomrule
\end{tabular}
\caption{Examples of \uline{terms} and \dashuline{definitions} from 
\citet{peters-etal-2018-deep}. Each row shows a \uline{term} (e.g., $s^{task}$)  along with its  \dashuline{definition} (e.g., ``softmax-normalized weights'').}
\label{tab:elmo-examples}
\end{table}

In formal terms, definition detection is comprised of  two tasks: classifying sentences as containing definitions or not, and identifying which spans within these sentences contain terms and definitions. As the performance of definition extractors continues to improve, these algorithms could pave the way for new types of intelligent assistance for readers of dense technical documents. For example, one could envision future interfaces that reveal definitions of  jargon like ``biLM'' or the symbol ``$s^{task}$'' when a reader hovers over the terms in a reading application \cite{head2020augmenting}.
Examples of sentences containing terms and definitions are shown in Table~\ref{tab:elmo-examples}.

Despite recent advances in definition detection, much work   remains   to be done before models are capable of extracting definitions with an accuracy appropriate for real-world applications. The first challenge is one of recall: existing systems are typically not trained to identify \textit{all} definitions in a document, but rather to classify individual sentences arbitrarily sampled from a large corpus. The second challenge is one of precision: the state of the art misclassifies upwards of 30\% of sentences \cite{Veyseh2020AJM}. This begs the questions of why definition extractors fall short, and how these shortcomings can be overcome.

In this paper, we contribute the following:

\begin{itemize} 
    
    \item An in-depth error analysis of the current best-performing model. This analysis characterizes the state of the field and illustrates future directions for improvement;
    
    \item A new model, Heuristically-Enhanced Deep Definition Extraction (\method{}), that extends a state-of-the-art model with improvements designed to address the problems found in the error analysis. An evaluation shows that this improved model outperforms the state of the art by a large margin (+12.7 F1);
    
    \item An introduction of the challenging task of full-document definition detection. In this task, models are evaluated based on their ability to identify definitions across an entire document's sentences. We believe this framing of definition detection is critical to preparing future algorithms for real-world use;
    
    \item A preliminary analysis of previous models and our model on the document-level definition detection task using a small test set of scholarly papers where every term and definition has been labeled. This analysis shows that \method{} outperforms the state of the art, while revealing opportunities for future improvements.
    
\end{itemize}

In summary, this paper draws attention to the work yet to be done in addressing the task of document-level definition detection for scholarly documents. We draw attention to the fact that a seemingly straightforward task like definition detection still poses significant challenges to NLP, and that this is an area that needs more focus in the scholarly document processing community.

\section{Related Work}\label{sec:related_work}

Definition detection has been tackled in several ways in prior research.
The traditional rule-based systems \cite{Muresan02amethod,westerhout-monachesi-2008-creating,westerhout2009definition} used hand-written definition patterns (e.g., ``is defined as``) and linguistic features (e.g., pronoun, verb, punctuation), providing high precision but low recall detection.
To address the low recall problem, model-driven approaches \cite{fahmi-bouma-2006-learning,westerhout-2009-extraction,navigli-velardi-2010-learning,reiplinger-etal-2012-extracting} were developed using statistical and syntactic features such as bag-of-words, sentence position, part-of-speech (POS) tags, and their combination with  hand-written rules. 
Notably, \citet{jin-etal-2013-mining} used conditional random field (CRF) \cite{lafferty2001conditional} to predict tags of each token in a sentence such as \texttt{TERM} for term tokens, \texttt{DEF} for definition tokens, and \texttt{O} for neither.
Recently, sophisticated neural models  such as convolutional networks \cite{espinosa-anke-schockaert-2018-syntactically} and graph convolutional networks \cite{Veyseh2020AJM} have been applied to obtain better sentence representations in combination with  syntactic features. However, our analysis found that the state-of-the-art is still far from solving the problem, achieving an F1 score of only 60 points on a standard test set.

\begin{table*}[h]
\vspace{0mm}
\small
\begin{tabular}{@{}p{2.9in}p{1.2in}p{1.2in} p{.6in}@{}}
\toprule
\textbf{Sentences}  & \textbf{Cause} & \textbf{Patterns} & \textbf{Solutions}\\ 
\midrule
\textbf{[Equal]} is open in something of type collection where that collection is a \textit{\{partition of something\}} .  &
\raggedright $\bullet$ Overgeneralization: technical term bias \\ $\bullet$ description (is)  & \raggedright (\textit{none \linebreak  applicable})  & ? \\
\\
\hdashline[0.4pt/2pt]
\\    
A \uline{\textbf{[graph - based operator]}} defines \dashuline{a transformation on a multi-document graph (MDG) G which preserves \textit{ \{some of its properties while reducing the number\}}}\ldots{}  &
\raggedright $\bullet$ Complicated sentence structure & \textit{<term>} defines \textit{<def>}  & parsing features \\
\\\hdashline[0.4pt/2pt]\\
The \uline{Inductive Logic Programming learning method} that we have developed enables us to automatically \dashuline{extract from a corpus N - V pairs whose elements axe linked by one of the semantic relations defined in the qualia structure }... &
\raggedright $\bullet$ Unfamiliar or unseen vocabulary \linebreak
$\bullet$ Unseen patterns  & \raggedright \textit{<term>} that we have developed enables us to automatically \textit{<def>}  & generalize \linebreak patterns \\ 
\bottomrule
\end{tabular}
\caption{\label{tab:error_annotation_sentencen} Sample annotations from our analysis of errors produced by \citepos{Veyseh2020AJM} joint model when extracting definitions from the \texttt{W00} \cite{jin-etal-2013-mining} dataset. Each row includes a sentence annotated with gold labels for \uline{terms} and \dashuline{definitions}, and the system's predictions for [\textbf{terms}] and \{\textit{definitions}\} (``Sentences'').  Also shown is a class of error (``Cause''), surface patterns that we anticipate could be used to correct the detection of the definition (``Patterns''), and classes of improvements to make to the model (``Solutions''). The first row is an example of a false positive; the second row is a partially-correct prediction; and the third row is a false negative.  A transcription error (`axe' instead of `are') is retained from the dataset.
}
\end{table*}

\section{Error Analysis of the Leading System}\label{sec:sentence_error_analysis}
In order to inform our efforts to develop a more advanced system, we performed an in-depth error analysis of the results of the current leading approach to definition and term identification, the joint model by \citet{Veyseh2020AJM}. We analyzed the models' predictions on the  \texttt{W00} dataset \cite{jin-etal-2013-mining} since it matches our target domain of scholarly papers and is the dataset that the joint model was evaluated on. Of the 224 test sentences,\footnote{Note that we use the test set of \texttt{W00} for manual analysis, which is only 10\% of the entire dataset. In our experiment in \S\ref{sec:sentence_experiment}, we didn't use the test set used in this error analysis, but did cross-validation using the train set, following the experimental setup in \citet{Veyseh2020AJM}.} the \citet{Veyseh2020AJM} system got 111 correct. The first author annotated the   remaining 113  sentences for which the algorithm was partially or fully incorrect to ascertain the root causes of the errors.

We discovered four (for terms) and five (for definitions) major  causes for the erroneous predictions, as summarized in Table \ref{tab:cause_types}.
We illustrate three examples  in Table \ref{tab:error_annotation_sentencen}.
For each example, we also labeled surface patterns between the term and  definition (e.g., ``\textit{<term>} defines \textit{<def>}''), and potential algorithm improvements to address the underlying problem.

For instance, in the bottom-most example in Table~\ref{tab:error_annotation_sentencen}, the system did not predict any term or definition, although the sentence includes the term ``\uline{Inductive Logic Programming Learning method}'' and the definition ``\dashuline{extract from a corpus...}''. 
Our conjecture is that the underlying surface pattern is unseen in the training set and too complicated to be generalized; we annotate a potential solution as \textit{pattern generalization}. 

\begin{table}[h]
\vspace{0mm}\small\centering
\begin{tabular}{@{}rr@{}}
\toprule
\textbf{Top hypothesized causes of error} & \textbf{(\%)} \\
\midrule
Overgeneralization: technical term bias	&	48.6\%\\
Unfamiliar or unseen vocabulary	&	25.7\%\\
Complicated sentence structure	&	12.9\%\\
Entity detection	&	4.3\%\\
\midrule
Overgeneralization: technical term bias	&	28.9\%\\
Overgeneralization: surface pattern bias	&	23.3\%\\
Unseen patterns	&	14.4\%\\
Complicated sentence structure	&	12.2\%\\
Overgeneralization: description	&	3.3\%\\
\bottomrule
\end{tabular}
\caption{\label{tab:cause_types} Top causes of errors for  terms (top) and definitions (bottom)
}
\end{table}

We  rank the causes of errors by  frequency and summarize the results  in Table \ref{tab:cause_types}. 
For detection of  terms, nearly half of the error cases fall into  overgeneralization of technical terms: overly predicting words like ``equal'' and ``model'' as terms (e.g., the top example in Table \ref{tab:error_annotation_sentencen}).

\begin{table}[h]
\vspace{0mm}\small\centering
\begin{tabular}{@{}rr@{}}
\toprule
\textbf{Proposed error correction solution types} & \textbf{(\%)} \\
\midrule
Syntactic (POS, parse tree, entity, acronym)   &	29.2\%\\
Heuristics&	23.6\%\\
Better encoder/tokenizer, UNK &	18.0\%\\
Rules (surface patterns)&	11.3\%\\
Annotations* &	9.4\%\\
Pattern generalization & 5.7\% \\
Mathematical symbol detection & 1.9\% \\
More context & 0.9\% \\
\bottomrule
\end{tabular}
\caption{\label{tab:solution_types} Proportions of proposed error correction solution types.  Annotations* indicates  extremely ambiguous cases even for humans, so additional human annotations are required to disambiguate them.
}
\end{table}

We again rank the error correction solutions by frequency (Table \ref{tab:solution_types}).
We predict that $29\%$ of errors can be fixed by informing the system about syntactic features of the sentence such as part-of-speech tags, parse tree annotations, entities, or acronyms for more accurate detection. 
Surprisingly, simple heuristics (e.g., stitching up discontiguous token spans) seem likely to be highly effective to address the errors in Table \ref{tab:cause_types}, such as discarding output that does not successfully predict both a term and a definition.  
In the next section, we implement the first three solution types on top of the state-of-the-art system and report the resulting performance improvements.

\begin{figure}[h]
\centering
{
\includegraphics[width=.99\linewidth]{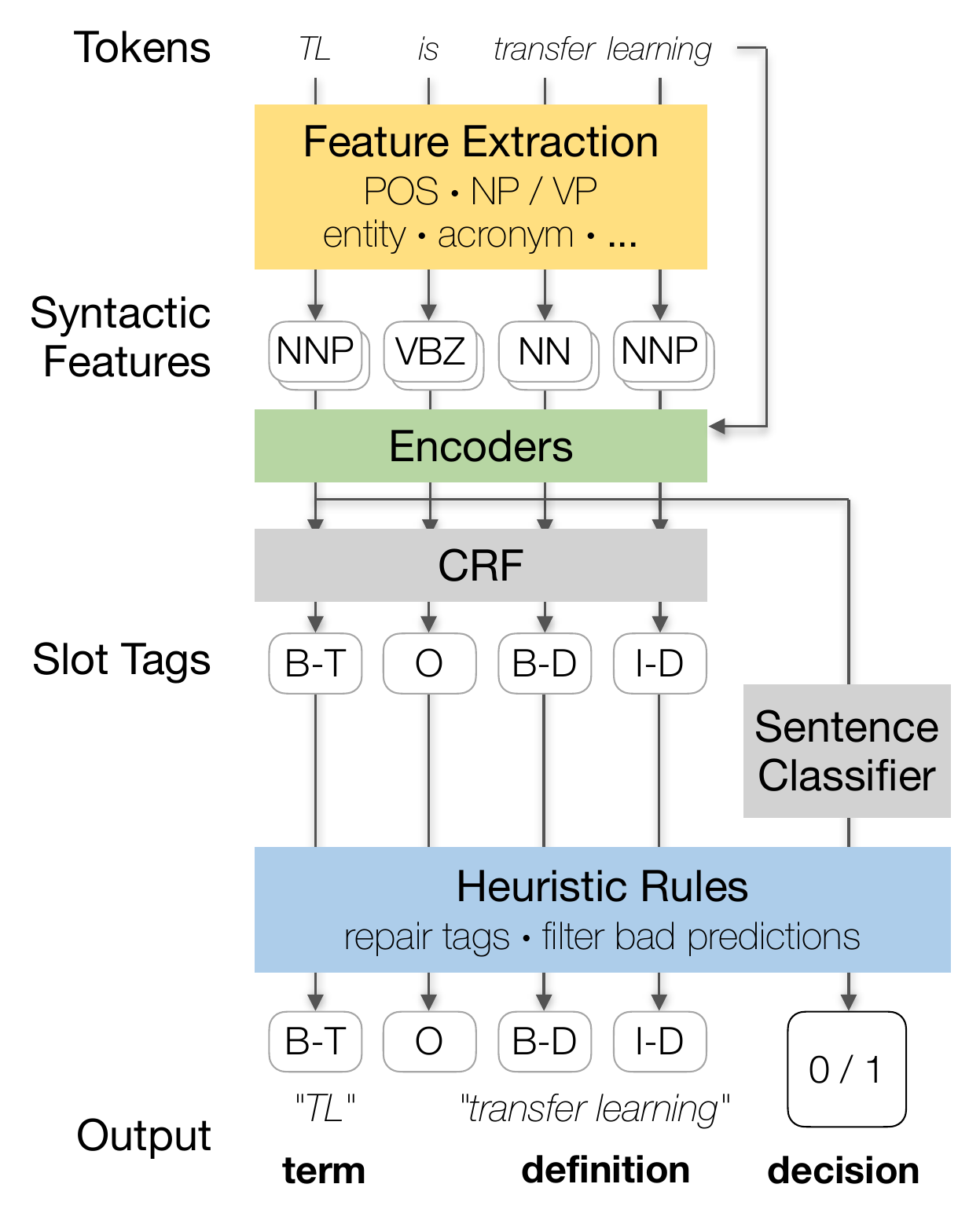}
 }
\caption{\label{fig:architecture} The \method{} model. The new modules developed in this work include the incorporation of syntactic features, the addition of pre-trained transformer encoders, and post-processing with heuristics.
}
\end{figure}

\section{Definition Sentence Detection Model}\label{sec:sentence_experiment}

To address the errors identified in \S\ref{sec:sentence_error_analysis}, we designed \method{}, a new sentence-level definition detection model. The model incorporates a set of syntactic features, heuristic filters, and encoders. Each of these  was designed to address a common class of error revealed in the error analysis. The model achieves superior performance over the state of the art for the task of sentence-level definition detection.

\subsection{Proposed Model: Heuristically-Enhanced Deep Definition Extraction (\method)}

\method{} extends the joint model proposed by \citet{Veyseh2020AJM}. 
The joint model is comprised of two components. The first component is a CRF-based sequence prediction model for slot tagging. The model assigns each token in a sentence one of five tags: term (``\texttt{B}-\texttt{TERM}'', ``\texttt{I}-\texttt{TERM}''), definition (``\texttt{B}-\texttt{DEF}'',``\texttt{I}-\texttt{DEF}''), or other (``\texttt{O}''). The second component is a binary classifier that labels each sentence as containing a definition or not.

\method{} has three new modules (Figure~\ref{fig:architecture}).
First, it encodes input from a transformer encoder fine-tuned on the task of definition extraction, whereas the joint model encodes input from a combination of a graph convolutional network and a BERT encoder without  fine-tuning.\footnote{However, we note that we were unable to replicate the accuracy reported by \citet{Veyseh2020AJM} when using the code provided by the authors.}
We evaluate several state-of-the-art encoders for this task, including BERT \cite{devlin2018bert}, RoBERTa \cite{liu2019roberta}, and SciBERT \cite{beltagy2019scibert}.

Second, \method{} is provided with additional syntactic features as input. These features include parts of speech, syntactic dependencies, and the token-level labels provided by entity recognizers and abbreviation detectors  \cite{Schwartz2003ASA}. The features are extracted using off-the-shelf tools like Spacy\footnote{\url{https://spacy.io/}} and SciSpacy \cite{neumann2019scispacy}.

Third, the output of the CRF and sentence classifier is refined using heuristic rules. The rules clean up the slot tags produced by the CRF, and override predictions made by the sentence classifier. The rules include, among other rules:

\begin{itemize} 
    \item Do not classify a sentence as a definition if it only contains a term without a definition, or a definition without a term.
    \item Stitch up discontiguous token spans for terms and definitions by assigning all contiguous tokens between two term or definition labels the same label.
\end{itemize}

These three enhancements developed in \method were selected specifically to suit the shortcomings of the models identified in the error analysis (\S\ref{sec:sentence_error_analysis}), leading to significant improvements for definition detection in our experiments.

\subsection{Baseline Models}

To evaluate the impact of these improvements on the definition detection task, \method{} was compared to four baseline systems: (1) DefMiner \cite{jin-etal-2013-mining}, a CRF-based sequence prediction model that makes use of hand-written features; (2) \citepos{Li2016DefinitionEW} model comprised of a CRF with an LSTM encoder; (3) GCDT \cite{liu-etal-2019-gcdt}, a global and local context encoder; (4) \citepos{Veyseh2020AJM} joint model described above. The experimental setup for the models followed the setup described by \citet{Veyseh2020AJM}.

\subsection{Metrics}\label{sec:metrics}

The models were compared using a set of metrics for both slot tagging and sentence classification on the \texttt{W00} test set. To evaluate the slot tagger, macro-averaged precision, recall, and F1 score were measured (column ``Macro P/R/F'' in Table~\ref{tab:result_sentence_level}). 
However, the Macro scores do not show performance specific to terms or definitions. 
Also, macro-averaging over the position tags (\texttt{B}, \texttt{I}) makes it  difficult to interpret general performance.
Therefore, we measured these three metrics only for term tags (``TERM P/R/F''); \texttt{B}-\texttt{TERM} and \texttt{I}-\texttt{TERM}, and definition tags (``DEF P/R/F''); \texttt{B}-\texttt{DEF} and \texttt{I}-\texttt{DEF}.
To evaluate the precision of the bounds of term and definition spans, we also evaluated the degree of overlap between each detected term or definition span and the corresponding span in the gold dataset (``Partial F'').
Furthermore, the accuracy of sentence classification was measured (column ``Classification'').
For each of these metrics, a higher score indicated superior performance.
We averaged  each score across 10-fold cross validation.

\subsection{Setup}

Due to  computing limitations, we chose the best hyper-parameter set through parameter sweeping with \method with the BERT encoder only, and use the best hyperparameters for all other models.
Here is the ranges of each parameter we tuned: batch sizes in \{8, 16, 32\}, number of training epochs [30, 50, 100] maximum length of sentences in \{80, 256, 512\}, learning rates in [\{2,5\}e-\{4,5\}].

The other parameters used as defaults in our experiments were as follows: the dropout ratio was 10\%, the layer size for POS embeddings was 50, and the hidden size for slot prediction was 512. 
We follow the default hyper-parameters for each transformer model of each size (base or large) using  HuggingFace's transformer libraries.\footnote{\url{https://github.com/huggingface/transformers}}

\begin{table*}[h]
\vspace{0mm}
\small\centering
\begin{tabular}{@{} l c c c cc @{}}
\toprule
& \textbf{Macro P/R/F}  & \textbf{TERM P/R/F} & \textbf{DEF P/R/F}  & \textbf{Partial F} & \textbf{Clsf.} \\ 
\midrule
DefMiner \cite{jin-etal-2013-mining}    &   52.5    / 49.5  / 50.5 & - & - & - & -\\
LSTM-CRF \cite{Li2016DefinitionEW}      &57.1       / 55.9  / 56.2 & - & - & - & -\\
GCDT \cite{liu-etal-2019-gcdt}          & 57.9      / 56.6  / 57.4 & - & - & - & -\\
Joint \cite{Veyseh2020AJM}             & 60.9     / 60.3 / 60.6 & - & - & - & -\\
Joint* \cite{Veyseh2020AJM}                              & 61.0 / 60.2 / 60.7     & - & -&-& 70.5 \\
\midrule
\method{}\\
\,\, Joint* + BERT-base                       & 59.5 / 61.3 / 60.3 & 66.6 / 70.0 / 68.2 & 72.1 / 74.0 / 72.8 & 74.3 &  83.4 \\
\,\, Joint* + BERT-large                      & 60.4 / 61.4 / 60.7     & 67.5 / 71.0 / 69.0 & 72.3 / 73.9 / 72.9 & 74.5 & 83.2\\
\,\, Joint* + RoBERTa-large                   & 60.3 / 61.6 / 60.7     & 67.3 / 70.3 / 68.6 & 72.8 / \textbf{74.6} / 73.5 & 73.2 & 84.2\\
\,\, Joint* + SciBERT                         & 61.9 / 61.2 / 61.5  & \textbf{71.1} / 69.1 / 69.9 & 74.0 / \textbf{74.6} / \textbf{74.2} & \textbf{75.7} & \textbf{85.1}\\
\hdashline[0.4pt/2pt]
\,\, Joint* + SciBERT + Syntactic         &   61.6 / 61.8 / 61.6  & 70.7 / 71.3 / \textbf{70.9} & 73.3 / 72.4 / 72.8 & 74.3  & 84.3\\ 
\hdashline[0.4pt/2pt]
\,\, Joint* + SciBERT + Syntactic + Heuristic          &  \textbf{72.9} / \textbf{74.3} / \textbf{73.4} & 69.8 / \textbf{72.1} / 70.8 & \textbf{75.4} / 71.8 / 73.3 & 74.3 & 84.5\\ 
\bottomrule
\end{tabular}
\caption{\label{tab:result_sentence_level} Comparison of the accuracy of recent models and the \method{} model for the definition detection task. Asterisks (*) indicate that we reimplemented the model from the authors' specification. Models were evaluated on the \texttt{W00} \cite{jin-etal-2013-mining} test set. Each score is averaged across 10-fold cross validation. 
\
Accuracy measurements include \textbf{P}recision, \textbf{R}ecall, and \textbf{F}1-score. Each of these measurements is macro-averaged. 
}
\end{table*}

\subsection{Results}

Outcomes of the evaluation for all measurements are presented in Table~\ref{tab:result_sentence_level}. 
The pre-trained language model encoders (BERT, RoBERTa, SciBERT) achieve comparable performance to more complex neural architectures like the graph convolutional networks used in \citepos{Veyseh2020AJM}. Models that included SciBERT \cite{beltagy2019scibert}, rather than BERT or ROBERTa, achieved higher accuracy on most measurements. We attribute this to the domain similarity between the scholarly documents that SciBERT was trained on, and those used in our evaluation.

With SciBERT as the base encoder, the incorporation of syntactic features led to further accuracy gains. Of particular note are the improvements in recall in term spans. During our evaluation, we observed that the gains from syntactic features were more pronounced for encoders with a small mode size (i.e., the ``-base'' models). We conjecture that this is because the larger encoder models were capable of learning comparable linguistic patterns to those captured by the syntactic features.

The addition of heuristic rules led to significant improvement (+11.8 Macro F1) over the combination of Joint and SciBERT. Given the modest improvement in term and definition tagging, we suspect that much of this improvement can be accounted for by  the correction of position markers in the slot tags (i.e., distinguishing between \texttt{B} and \texttt{I} in the tag assignments).

In the following experiments, we call \method  the combination of  three components: the encoder (SciBERT or RoBERTa), syntactic features, and heuristic filters.

\section{Document-Level Definition Detection}\label{sec:document_def_extraction}
Although HEDDEx attains reasonable performance on individual sentences, it faces new challenges when applied to the  scenario of document-level analysis.
In this section, we evaluate sentence detection for full papers in two novel ways.  First, we assess the \textit{precision} of the  \method model across \textit{all of the sentences} of 50 documents in \S\ref{sec:error_analysis_on_predicted_definitions}.  Second, we assess both the \textit{precision} and the \textit{recall} of the algorithm across \textit{all of the sentences} of 2 full documents (\S\ref{sec:full_document_defniition_annotation}, S\ref{sec:docdef_document_level_detection}).

\subsection{Error Analysis on Predicted Definitions }\label{sec:error_analysis_on_predicted_definitions}

To assess how well \method{} works at the document level, we randomly sampled
50 ACL papers from the S2ORC dataset  \cite{lo-wang-2020-s2orc},  a large corpus of 81.1M English-language academic papers spanning many academic disciplines.   We ran the pretrained \method model on every sentence of every document; if the model detected a term/definition pair, the corresponding sentence was output for assessment.  (Note that this analysis can estimate precision but not recall, as false negatives are not detected.)

We replace all citations and references to figures, tables, and sections with corresponding placeholders (e.g., CITATION, FIGURE), but keep raw \TeX{} format of mathematical symbols in order  to retain the   structure of the equations. 
From the 50 ACL papers, the model detected $924$ definitions out of $13,658$ sentences and the average number of definitions per paper is $18.5$.

\begin{table}[h]
\vspace{0mm}
\small\centering
\begin{tabular}{@{}r@{\hskip 2mm}r@{\hskip 2mm}r@{\hskip 2mm}r@{}}
\toprule
\multicolumn{2}{@{}r}{\textbf{Term (\%)}}  & \multicolumn{2}{@{}r}{\textbf{Definition (\%)}}\\
\midrule
	Textual term & 	45.2\%          & Textual Def.	&	58.7\%\\
	Incorrect term & 	27.3\%      & Other: implausible	&	24.8\%\\
	Math symbol term        & 	22.7\% & Other: plausible	&	11.8\%\\
	Acronym  & 	3.3\%  & Short name / Synonym	&	3.5\%\\
	Acronym and text & 	1.3\%          & Textual \& Formula Def.	&	0.6\%\\
	&  & Formula Def.	&	0.4\% \\
\bottomrule
\end{tabular}
\caption{\label{tab:error_analysis_prediction} Analysis of \method{} output on 50 ACL papers, ordered by frequency.  N = 923.
}
\end{table}

The third author evaluated the predicted terms and definitions separately by choosing one among the labels  shown in Table \ref{tab:error_analysis_prediction}.
For terms, the algorithm correctly labeled $72.5\%$. We subdivide these correctly labeled terms into standard terms (45.2\%), math symbols (22.7\%), acronyms, acronym (3.3\%), or acronym and text (1.3\%).   Among the correctly labeled definitions (total 63.2\% = 58.7\%+3.5\%+0.6\%+0.4\%), 92.6\%  are textual definitions, 5.6\% are short names or synonyms, and 1.7\% include mathematical symbols.   We divided non-definitional text into two types: plausible ($24.8\%$) and implausible ($11.8\%$), which signals an error.  The plausible text refers to explanations or secondary information (similar to DEFT \cite{spala-etal-2019-deft}'s secondary definition, but without sentence crossings).

\begin{table}[h]
\vspace{0mm}
\small\centering
\begin{tabular}{@{}r@{\hskip 1.8mm}r@{\hskip 1.8mm}r@{\hskip 2mm}r@{}}
\toprule
\multicolumn{2}{@{}r}{\textbf{Term Span (\%)}}  & \multicolumn{2}{@{}r}{\textbf{Definition Span (\%)}}\\
\midrule
	Correct	&	83.4\% & Correct	&	89.9\%\\
	Too Long (to the right)	&	10.1\% & Cut Off (to the right)	&	3.4\%\\
	Cut Off (to the right)	&	3.2\% & Too Long (to the left)	&	2.6\%\\
	Cut Off (to the left)	&	1.6\% & Cut Off (to the left)	&	2.4\%\\
	Too Long (to the left)	&	1.4\% & Too Long (to the right)	&	1.3\% \\
\bottomrule
\end{tabular}
\caption{\label{tab:error_analysis_prediction_span}
Analysis of span length of \method{} output on 50 ACL papers, ordered by frequency. $N$ = 923.
}
\end{table} 

We also measured whether the predicted span length is correct, too long, or cut off (Table \ref{tab:error_analysis_prediction_span}).  These scores are quite high; 
 $83.4\%$ correct for terms and $89.9\%$ for definitions (see Table \ref{tab:document_prediction_elmo}).

\subsection{Full Document Definition Annotation}\label{sec:full_document_defniition_annotation}

Prior definition annotation collections select unrelated  sentences from across a document collection.  As mentioned in the introduction, we are interested in annotating full papers, which requires finding \textit{every} definition within a given paper. 
Therefore, we created a new collection in which we annotate every sentence within a document, allowing assessment of  recall as well as precision. 
Two annotators annotated two full papers using an  annotation scheme similar to that used in DEFT \cite{spala-etal-2019-deft} except for omitting cross-sentence links.

We chose to annotate two award-winning ACL papers: ELMo \cite{peters-etal-2018-deep} and LISA \cite{strubell-etal-2018-linguistically} resulting in 485 total sentences from which we identified 98 definitional and 387 non-definitional sentences.
Similar to DEFT \cite{spala-etal-2019-deft}, we measured inter-annotator agreement using Krippendorff's alpha \cite{krippendorff2011computing} with the MASI distance metric \cite{passonneau2006measuring}.
We obtained 0.626 for terms and 0.527 for definitions, where the agreement score for terms is lower than those in DEFT annotations (0.80).
This may be because our annotations for terms include various types such as textual terms, acronyms, and math symbols, while terms in DEFT are only textual terms.
The task was quite difficult: each annotator takes two and half hours to annotate a single paper.
Future work will include refining the annotation scheme to ensure more consistency among annotators and to annotate more documents.

\subsection{Evaluation on Document-level Definitions}\label{sec:docdef_document_level_detection}

We evaluated document-level performance using the same metrics used in \S\ref{sec:metrics}.
All metrics were averaged over scores from 10-fold validation models.
The ensemble model aggregates ten system predictions from the 10-fold validation models and choose the final label via majority voting. 
We use the best single system; \method but with RoBERTa,\footnote{RoBERTa and SciBERT show comparable performance on the document-level definition detection task.} for model ensembling.

\begin{table}[h]
\vspace{0mm}
\small\centering
\begin{tabular}{@{}p{3.1cm}@{\hskip 2mm}p{0.55cm}p{0.55cm}@{\hskip 5.5mm}p{0.55cm}@{\hskip 2.5mm}p{0.55cm}@{\hskip 5.5mm}p{0.5cm}@{}}
\toprule
& \textbf{Macro} & \textbf{TERM} & \textbf{DEF} & \textbf{Partial} & \textbf{Clf.}\\
\midrule
Joint model	& 36.0 & 30.1 & 38.1 & 34.1 & 86.8\\
\hdashline[0.4pt/2pt]
\method w/ BERT& 45.3 & 32.4 & 39.0 & 34.6 & 89.1\\
\method w/ RoBERTa& 47.7 & 36.4 & 47.2& 37.2& 88.1\\
\hdashline[0.4pt/2pt]
\method ensemble	&	\textbf{50.4} & \textbf{38.7} &\textbf{49.5} &\textbf{39.0} & \textbf{89.8}\\
\bottomrule
\end{tabular}
\caption{\label{tab:document_level_evaluation} Document-level evaluation on our annotated  documents. F1 score is measured for every metric except for classification (Clf.), which uses accuracy.  
}
\end{table}

Compared to the joint model by \citet{Veyseh2020AJM}, \method showed significant improvements on every evaluation metric, which is slightly larger than that of the sentence-level evaluation (Table \ref{tab:document_level_evaluation}).
With model ensembling, compared to the state-of-the-art system, \method achieved gains by +14.4 Macro F1 points, +8.7 TERM F1 points, +11.4 DEF F1 points, +4.9 Partial Matching F1 points, and +3.0 classification accuracy scores.

\begin{table}[h]
\vspace{0mm}
\small\centering
\begin{tabular}{@{}l ccc@{}}
\toprule
 & \textbf{Precision} & \textbf{Recall}  & \textbf{F1}\\
\midrule
Macro   &55.3&	46.7&	50.4\\
TERM & 44.8	 & 34.0 &	38.7\\
DEF & 55.6 & 44.7 & 49.5\\
\bottomrule
\end{tabular}
\caption{\label{tab:best_system_with_prf} Low recall problem in document-level definition detection. We report precision, recall, and f1 scores on three metrics; Macro, TERM, and DEF, using our best system; \method ensemble.
}
\end{table}

However, document-level definition detection is a much harder task than sentence-level detection.
Compared to the sentence-level task in Table \ref{tab:result_sentence_level}, the document-level task showed relatively lower performance (73.4 Macro F1 in sentence-level versus 50.4 Macro F1 in document-level).
In particular, recall is much lower than precision in the document-level task (Table \ref{tab:best_system_with_prf}), whereas in the sentence-level task, precision and recall are almost the same, indicating the necessity of incorporation of document structure as additional features (See further discussion in \S\ref{sec:discussion}).

\begin{table*}[h!]
\small\centering
\begin{tabular}{@{}p{.03in}p{5.6in}p{.34in}@{}}\\
\toprule
& \textbf{Predicted definition sentences} & \textbf{Type} \\ 
\midrule
1 & Our \uline{word vectors} are \dashuline{learned functions of the internal states of a \textit{\{deep bidirectional language model\}} (\textbf{[biLM]}), which is pre-trained on a large text corpus}.		&	term \\\midrule 
2 & We use vectors derived from a bidirectional LSTM that is trained with a coupled \dashuline{\textit{\{language model\}}} (\uline{\textbf{[LM]}}) objective on a large text corpus.	&	term \\\midrule 
3 & Using intrinsic evaluations, we show that the higher-level \textbf{[LSTM states]} capture context-dependent aspects of word meaning (e.g., they can be used without modification to perform well on supervised word sense disambiguation tasks) while lower-level states \textit{\{model aspects of syntax\}}.		&	term \\\midrule 
4 & We first show that they can be easily added to existing models for six diverse and challenging \uline{language understanding problems} , including \dashuline{textual entailment, question answering and sentiment analysis}. & term \\\midrule
5 & For tasks where direct comparisons are possible,  outperforms \uline{\textbf{[CoVe]}} CITATION, which \dashuline{\textit{\{computes contextualized representations using a neural machine translation encoder\}}}.	&	term \\\midrule 
6 & \uline{context2vec} CITATION \dashuline{uses a bidirectional Long Short Term Memory LSTM ; CITATION to encode the context around a pivot word} . & term \\\midrule
7 & Unlike most widely used word embeddings CITATION, \uline{word representations} are \dashuline{functions of the entire input sentence}, as described in this section. & term \\\midrule
8 & This setup allows us to do \textbf{[semi-supervised learning]}, where the biLM is pretr\textit{\{ained at a large scale\}} (Sec. SECTION) and easily incorporated into a wide range of existing neural NLP architectures (Sec. SECTION).		&	term \\\midrule 
9 & Given a sequence of $N$ tokens, $(t_1, t_2, ..., t_N)$, a \uline{forward language model} \dashuline{computes the probability of the sequence by modeling the probability of token $t_k$ given the history $(t_1, ..., t_{k-1})$:} & term \\\midrule
10 & A \uline{\textbf{[backward LM]}} is \textit{\{\dashuline{similar to a forward LM, except it runs over the sequence in reverse, predicting the previous token given the future context}\}}:	&	term \\\midrule 
11 & A \textbf{[biLM]} \textit{\{combines both a forward and backward LM\}}.		&	term \\\midrule 
12 & where \textbf{[$\pmb{h^{LM}_{k,0}}$]} is \textit{\{the token layer\}} and ${h}^{LM}_{k,j} = [\overrightarrow{{h}}^{LM}_{k,j}; \overleftarrow{{h}}^{LM}_{k,j}]$, for each biLSTM layer.	&	symbol \\\midrule 
13 & In (EQUATION), \textbf{[$s^{task}$]} are \textit{\{softmax-normalized weights\}} and the scalar parameter $\gamma^{task}$ allows the task model to scale the entire  vector.		&	symbol \\\midrule 
14 & For each token $t_k$, a \uline{$L$-layer biLM} \dashuline{computes a set of $2L + 1$ representations EQUATION where $h^{LM}_{k,0}$ is the token layer and ${h}^{LM}_{k,j} = [\overrightarrow{{h}}^{LM}_{k,j}; \overleftarrow{{h}}^{LM}_{k,j}]$, for each biLSTM layer.}& symbol, term \\\midrule
15 & In (EQUATION), \uline{$s^{task}$} are \dashuline{softmax-normalized weights} and the \textbf{[\uline{scalar parameter $\gamma^{task}$}]} \textit{\{\dashuline{allows the task model to scale the entire  vector}\}}.	&	symbol \\\midrule 
16 & \uline{The} \textbf{[\uline{Stanford Question Answering Dataset (SQuAD)} CITATION]} \textit{\{\dashuline{contains 100K+ crowd sourced question-answer pairs where the answer is a span in a given Wikipedia paragraph}\}}.	&	term \\\midrule 

17 & \textbf{[\uline{Textual entailment}]} is \textit{\{\dashuline{the task of determining whether a ``hypothesis'' is true, given a ``premise''}\}}.		&	term \\\midrule 
18 & The \textbf{[Stanford Natural Language Inference (SNLI) corpus CITATION]} \textit{\{provides approximately 550K hypothesis/premise pairs\}}.	&	term \\\midrule 
19 & A \textbf{[\uline{semantic role labeling (SRL) system}]} \textit{\{\dashuline{models the predicate-argument structure of a sentence}\}}\dashuline{, and is often described as answering.}		&	term \\\midrule 
20 & CITATION modeled \textbf{[SRL]} \textit{\{as a BIO tagging problem and used an 8-layer deep biLSTM with forward and backward directions interleaved\}}, following CITATION.		&	term \\\midrule 
21 & \textbf{[\uline{Coreference resolution}]} is \textit{\{\dashuline{the task of clustering mentions in text that refer to the same underlying real world entities}\}}.		&	term \\\midrule 
22 & The \textbf{[CoNLL]} 2003 NER task CITATION \textit{\{consists of newswire from the Reuters RCV1 corpus tagged with four different entity types (PER, LOC, ORG, MISC)\}}.		&	term \\\midrule 
23 & The \textbf{[fine-grained sentiment classification]} task in the Stanford Sentiment Treebank SST-5 \textit{\{involves selecting one of five labels (from very negative to very positive) to describe a sentence from a movie review\}}.	&	term \\\midrule 
24 & The sentences contain diverse \uline{linguistic phenomena} such as \dashuline{idioms} and \uline{complex syntactic constructions} such as \dashuline{negations} that are difficult for models to learn. & multi-term \\\midrule
25 & Intuitively, the \textbf{[biLM]} must be \textit{\{disambiguating the meaning of words using their context\}}.	&	term \\\midrule 
26 & a fine grained \textit{\{word sense disambiguation\}} (\textbf{[WSD]}) task and a POS tagging task.&	term \\ 
\bottomrule
\end{tabular}
\caption{\label{tab:document_prediction_elmo} All predicted and gold label terms and definitions for the ELMo paper \cite{peters-etal-2018-deep}.
Gold labels for \uline{terms} are \uline{underlined} and for \dashuline{definitions} are \dashuline{dashed}.
System-predicted [\textbf{terms}] are placed in [\textbf{boldfaced brackets}] and \{\textit{definitions}\} are placed  in \{\textit{italicized braces}\}.  
``CITATION,'' ``SECTION,'' and ``EQUATION'' are placeholders inserted for citations, section numbers, and display equations. 
``Type'' means term type. 
}
\vspace{-10mm}
\end{table*}

Table \ref{tab:document_prediction_elmo} 
shows the predicted terms and definitions as well as annotated gold labels.  
Acronym patterns (e.g., ``biLM,'' ``WSD''), definition of newly-proposed terms (e.g., ``LISA''), re-definition of prior work (e.g., ``SQuAD,'' ``SRL,'' Coreference resolution) and some of mathematical symbols were detected well.
However, as  sentences get more complex, the system made incorrect predictions.
Additionally, sub-words or parentheses in abbreviations are sometimes partially predicted (e.g., the beginning of the word ``pretrained'' is cut off in the definition of ``semi-supervised learning'' in example 8 of Table 10)
.

However, the aforementioned problem of low recall is severe for this task, particularly since the model often fails to detect mathematical symbols or a combination of textual terms and mathematical symbols (e.g., ``\uline{$L$-layer biLM}''). Moreover, when a sentence contains multiple terms and/or multiple symbols together, the system only ever detects one of them.

\section{Discussion}\label{sec:discussion}
Detecting definitions is a very challenging task, and it is far from solved. 
Here we  discuss  remaining challenges and ideas for improvements, and motivate the need for high-precision, high-recall definition detection in an academic document reading aid application. 

Outstanding technical challenges include: 
\begin{itemize} 
    \item \textbf{Poor recognition of mathematical symbols}:
    As shown in our experiment, our system is less successful at  detecting math symbols  than textual terms.
    This is mainly because the lack of coverage of mathematical symbols in our training dataset (\texttt{W00}).

    \item \textbf{Contextual disambiguation of 
    symbols}: In our study, we observe that some symbols are used with multiple meanings. For example, symbol $T$ in the LISA paper is used for \textit{token representation} as well as \textit{matrix transpose}. Disambiguating terms based on context of use will be an interesting future direction. 
    \item \textbf{Description vs Definition}: 
    In our annotation and error analysis, the most difficult distinction was between 
    definitions and descriptions --- they have quite similar surface patterns, although they refer to entirely different meanings.  
    For instance, a definition is the exact denotation of a word, while a description is more detailed so it can change from person to person.
    Training a model that distinguishes these types should lead to better and more useful results.
\end{itemize}

Potential ideas for improvements of the system include:
\begin{itemize}[noitemsep,topsep=1pt]
   \item \textbf{Annotation of mathematical definitions}:
    A solution for  poor math symbol detection is to annotate math symbols and use them for our training. One option is to add span information to the binary judgements of the math definition collection of  \citet{vanetik-etal-2020-automated}.
    \item   \textbf{Utilization of document-level features}: 
    Document structure and positional information  may improve  detection.
    For instance, the section information of a term would be an important feature to recognize whether a term is first introduced or not.
    \item \textbf{Data augmentation or domain-specific fine-tuning for high-recall system}: 
    Existing definition training sets are small (\texttt{W00} contains only 731 definitional sentences).  To obtain more data, the data can be augmented via seed patterns or fine-tuning with  existing language models such as SciBERT. 
\end{itemize}

Lastly, as the performance of definition detection systems increases, these systems can be applied to real-world reading or writing aids.
We discuss potential issues of our system in the realistic settings:
\begin{itemize}[noitemsep,topsep=1pt]
    \item \textbf{Metrics for usefulness}: Currently, we measure precision, recall, and F1 scores with the document-level annotations. However, we have not explored the usefulness of the predicted definitions for readability, when they are used in real-world applications like ScholarPhi \cite{head2020augmenting}.  Deciding when and where to show definitions based on context and information density still remains an important future direction.
    \item \textbf{Categorization of definitions}: We observe that in fact, terms and definitions can be grouped into multiple categories: short names, acronyms, textual definitions, formula definitions,  and more. Automatically categorizing these and showing  structured definitions might be helpful for organizing and ranking definitions in a user interface. 
    \item \textbf{Repeated definitions and terms within documents}: 
    We  observed a pattern in which the same term is referred to multiple times in slightly different ways. 
    Newly proposed terms are especially likely to exhibit this pattern.
    Grouping  and summarizing these in a \textit{glossary table} would be helpful for an academic document reader application.
\end{itemize}

\section{Conclusion}\label{sec:conclusion}
This work sets the stage for bridging the gap between a well-known NLP task; \textit{definition detection}, and  real-world applications of the technique that requires both high precision and high recall. 
To achieve the goal, we proposed a more realistic setup for definition detection task called \textit{document-level definition detection} that requires high recall, mathematical symbol recognition, and document-level feature engineering.
Our proposed definition detection system \method achieved significant gains in both sentence-level and document-level tasks.
Yet, the problem is  far from being solved.  We suggest that better coverage of variability of expression, recognition of mathematical symbols and notation, and other nuances of the task must still be addressed.

\section*{Acknowledgements} 
We like to thank Amir Pouran Ben Veyseh for his help in sharing his code, preprocessed data, and general advice on replication of his work. 
We also thank Raymond Fok, Vivek Aithal, Hearst lab members at UC Berkeley and anonymous reviewers at SDP 2020 for their helpful comments.
This research receives funding from the Alfred P. Sloan
Foundation, the Allen Institute for AI, Office of Naval Research
grant N00014-15-1-2774,  NSF Convergence Accelerator
award 1936940, NSF RAPID award 2040196, and the University of Washington Washington Research Foundation/Thomas J. Cable Professorship.

\normalem
\bibliographystyle{acl_natbib}
\bibliography{main} 

\begin{thebibliography}{30}
\expandafter\ifx\csname natexlab\endcsname\relax\def\natexlab#1{#1}\fi

\bibitem[{Banerjee and Pedersen(2002)}]{banerjee2002adapted}
Satanjeev Banerjee and Ted Pedersen. 2002.
\newblock An adapted lesk algorithm for word sense disambiguation using
  wordnet.
\newblock In \emph{International conference on intelligent text processing and
  computational linguistics}, pages 136--145. Springer.

\bibitem[{Beltagy et~al.(2019)Beltagy, Lo, and Cohan}]{beltagy2019scibert}
Iz~Beltagy, Kyle Lo, and Arman Cohan. 2019.
\newblock Scibert: A pretrained language model for scientific text.
\newblock In \emph{Proceedings of the 2019 Conference on Empirical Methods in
  Natural Language Processing and the 9th International Joint Conference on
  Natural Language Processing (EMNLP-IJCNLP)}, pages 3606--3611.

\bibitem[{Bovi et~al.(2015)Bovi, Telesca, and Navigli}]{bovi2015large}
Claudio~Delli Bovi, Luca Telesca, and Roberto Navigli. 2015.
\newblock Large-scale information extraction from textual definitions through
  deep syntactic and semantic analysis.
\newblock \emph{Transactions of the Association for Computational Linguistics},
  3:529--543.

\bibitem[{Devlin et~al.(2019)Devlin, Chang, Lee, and
  Toutanova}]{devlin2018bert}
Jacob Devlin, Ming-Wei Chang, Kenton Lee, and Kristina Toutanova. 2019.
\newblock {BERT}: Pre-training of deep bidirectional transformers for language
  understanding.
\newblock In \emph{Proceedings of the 2019 Conference of the North {A}merican
  Chapter of the Association for Computational Linguistics: Human Language
  Technologies, Volume 1 (Long and Short Papers)}, pages 4171--4186,
  Minneapolis, Minnesota. Association for Computational Linguistics.

\bibitem[{Espinosa-Anke and
  Schockaert(2018)}]{espinosa-anke-schockaert-2018-syntactically}
Luis Espinosa-Anke and Steven Schockaert. 2018.
\newblock \href {https://doi.org/10.18653/v1/N18-2061} {Syntactically aware
  neural architectures for definition extraction}.
\newblock In \emph{Proceedings of the 2018 Conference of the North {A}merican
  Chapter of the Association for Computational Linguistics: Human Language
  Technologies, Volume 2 (Short Papers)}, pages 378--385, New Orleans,
  Louisiana. Association for Computational Linguistics.

\bibitem[{Fahmi and Bouma(2006)}]{fahmi-bouma-2006-learning}
Ismail Fahmi and Gosse Bouma. 2006.
\newblock \href {https://www.aclweb.org/anthology/W06-2609} {Learning to
  identify definitions using syntactic features}.
\newblock In \emph{Proceedings of the Workshop on Learning Structured
  Information in Natural Language Applications}.

\bibitem[{Hashimoto et~al.(2011)Hashimoto, Torisawa, De~Saeger, Kurohashi
  et~al.}]{hashimoto2011extracting}
Chikara Hashimoto, Kentaro Torisawa, Stijn De~Saeger, Sadao Kurohashi, et~al.
  2011.
\newblock Extracting paraphrases from definition sentences on the web.
\newblock In \emph{Proceedings of the 49th Annual Meeting of the Association
  for Computational Linguistics: Human Language Technologies}, pages
  1087--1097.

\bibitem[{Head et~al.(2020)Head, Lo, Kang, Fok, Skjonsberg, Weld, and
  Hearst}]{head2020augmenting}
Andrew Head, Kyle Lo, Dongyeop Kang, Raymond Fok, Sam Skjonsberg, Daniel~S.
  Weld, and Marti~A. Hearst. 2020.
\newblock Augmenting scientific papers with just-in-time, position-sensitive
  definitions of terms and symbols.
\newblock \emph{arXiv preprint arXiv:2009.14237}.

\bibitem[{Huang et~al.(2019)Huang, Sun, Qiu, and
  Huang}]{huang-etal-2019-glossbert}
Luyao Huang, Chi Sun, Xipeng Qiu, and Xuanjing Huang. 2019.
\newblock \href {https://doi.org/10.18653/v1/D19-1355} {{G}loss{BERT}: {BERT}
  for word sense disambiguation with gloss knowledge}.
\newblock In \emph{Proceedings of the 2019 Conference on Empirical Methods in
  Natural Language Processing and the 9th International Joint Conference on
  Natural Language Processing (EMNLP-IJCNLP)}, pages 3509--3514, Hong Kong,
  China. Association for Computational Linguistics.

\bibitem[{Jin et~al.(2013)Jin, Kan, Ng, and He}]{jin-etal-2013-mining}
Yiping Jin, Min-Yen Kan, Jun-Ping Ng, and Xiangnan He. 2013.
\newblock \href {https://www.aclweb.org/anthology/D13-1073} {Mining scientific
  terms and their definitions: A study of the {ACL} anthology}.
\newblock In \emph{Proceedings of the 2013 Conference on Empirical Methods in
  Natural Language Processing}, pages 780--790, Seattle, Washington, USA.
  Association for Computational Linguistics.

\bibitem[{Krippendorff(2011)}]{krippendorff2011computing}
Klaus Krippendorff. 2011.
\newblock Computing krippendorff's alpha-reliability.
\newblock Technical report, University of Pennsylvania.
\newblock Retrieved from https://repository.upenn.edu/asc\_papers/43/.

\bibitem[{Lafferty et~al.(2001)Lafferty, McCallum, and
  Pereira}]{lafferty2001conditional}
John~D. Lafferty, Andrew McCallum, and Fernando C.~N. Pereira. 2001.
\newblock \href {http://dl.acm.org/citation.cfm?id=645530.655813} {Conditional
  random fields: Probabilistic models for segmenting and labeling sequence
  data}.
\newblock In \emph{Proceedings of the Eighteenth International Conference on
  Machine Learning}, pages 282--289, San Francisco, CA, USA. Morgan Kaufmann
  Publishers Inc.

\bibitem[{Li et~al.(2016)Li, Xu, and Chung}]{Li2016DefinitionEW}
SiLiang Li, Bin Xu, and Tong~Lee Chung. 2016.
\newblock Definition extraction with lstm recurrent neural networks.
\newblock In \emph{Chinese Computational Linguistics and Natural Language
  Processing Based on Naturally Annotated Big Data}, pages 177--189. Springer.

\bibitem[{Liu et~al.(2019{\natexlab{a}})Liu, Meng, Zhang, Xu, Chen, and
  Zhou}]{liu-etal-2019-gcdt}
Yijin Liu, Fandong Meng, Jinchao Zhang, Jinan Xu, Yufeng Chen, and Jie Zhou.
  2019{\natexlab{a}}.
\newblock \href {https://doi.org/10.18653/v1/P19-1233} {{GCDT}: A global
  context enhanced deep transition architecture for sequence labeling}.
\newblock In \emph{Proceedings of the 57th Annual Meeting of the Association
  for Computational Linguistics}, pages 2431--2441, Florence, Italy.
  Association for Computational Linguistics.

\bibitem[{Liu et~al.(2019{\natexlab{b}})Liu, Ott, Goyal, Du, Joshi, Chen, Levy,
  Lewis, Zettlemoyer, and Stoyanov}]{liu2019roberta}
Yinhan Liu, Myle Ott, Naman Goyal, Jingfei Du, Mandar Joshi, Danqi Chen, Omer
  Levy, Mike Lewis, Luke Zettlemoyer, and Veselin Stoyanov. 2019{\natexlab{b}}.
\newblock Roberta: A robustly optimized bert pretraining approach.
\newblock \emph{arXiv preprint arXiv:1907.11692}.

\bibitem[{Lo et~al.(2020)Lo, Wang, Neumann, Kinney, and
  Weld}]{lo-wang-2020-s2orc}
Kyle Lo, Lucy~Lu Wang, Mark Neumann, Rodney Kinney, and Daniel Weld. 2020.
\newblock \href {https://doi.org/10.18653/v1/2020.acl-main.447} {{S}2{ORC}: The
  semantic scholar open research corpus}.
\newblock In \emph{Proceedings of the 58th Annual Meeting of the Association
  for Computational Linguistics}, pages 4969--4983, Online. Association for
  Computational Linguistics.

\bibitem[{Muresan and Klavans(2002)}]{Muresan02amethod}
A~Muresan and Judith Klavans. 2002.
\newblock A method for automatically building and evaluating dictionary
  resources.
\newblock In \emph{Proceedings of the Language Resources and Evaluation
  Conference (LREC}.

\bibitem[{Navigli and Velardi(2010)}]{navigli-velardi-2010-learning}
Roberto Navigli and Paola Velardi. 2010.
\newblock \href {https://www.aclweb.org/anthology/P10-1134} {Learning
  word-class lattices for definition and hypernym extraction}.
\newblock In \emph{Proceedings of the 48th Annual Meeting of the Association
  for Computational Linguistics}, pages 1318--1327, Uppsala, Sweden.
  Association for Computational Linguistics.

\bibitem[{Neumann et~al.(2019)Neumann, King, Beltagy, and
  Ammar}]{neumann2019scispacy}
Mark Neumann, Daniel King, Iz~Beltagy, and Waleed Ammar. 2019.
\newblock Scispacy: Fast and robust models for biomedical natural language
  processing.
\newblock In \emph{Proceedings of the 18th BioNLP Workshop and Shared Task},
  pages 319--327.

\bibitem[{Passonneau(2006)}]{passonneau2006measuring}
Rebecca~J Passonneau. 2006.
\newblock Measuring agreement on set-valued items (masi) for semantic and
  pragmatic annotation.
\newblock In \emph{Proceedings of the Fifth International Conference on
  Language Resources and Evaluation (LREC’06)}.

\bibitem[{Peters et~al.(2018)Peters, Neumann, Iyyer, Gardner, Clark, Lee, and
  Zettlemoyer}]{peters-etal-2018-deep}
Matthew Peters, Mark Neumann, Mohit Iyyer, Matt Gardner, Christopher Clark,
  Kenton Lee, and Luke Zettlemoyer. 2018.
\newblock Deep contextualized word representations.
\newblock In \emph{Proceedings of the 2018 Conference of the North {A}merican
  Chapter of the Association for Computational Linguistics: Human Language
  Technologies, Volume 1 (Long Papers)}, pages 2227--2237.

\bibitem[{Reiplinger et~al.(2012)Reiplinger, Sch{\"a}fer, and
  Wolska}]{reiplinger-etal-2012-extracting}
Melanie Reiplinger, Ulrich Sch{\"a}fer, and Magdalena Wolska. 2012.
\newblock \href {https://www.aclweb.org/anthology/W12-3206} {Extracting
  glossary sentences from scholarly articles: A comparative evaluation of
  pattern bootstrapping and deep analysis}.
\newblock In \emph{Proceedings of the {ACL}-2012 Special Workshop on
  Rediscovering 50 Years of Discoveries}, pages 55--65, Jeju Island, Korea.
  Association for Computational Linguistics.

\bibitem[{Schwartz and Hearst(2003)}]{Schwartz2003ASA}
Ariel~S. Schwartz and Marti~A. Hearst. 2003.
\newblock A simple algorithm for identifying abbreviation definitions in
  biomedical text.
\newblock \emph{Pacific Symposium on Biocomputing. Pacific Symposium on
  Biocomputing}, pages 451--62.

\bibitem[{Spala et~al.(2019)Spala, Miller, Yang, Dernoncourt, and
  Dockhorn}]{spala-etal-2019-deft}
Sasha Spala, Nicholas~A. Miller, Yiming Yang, Franck Dernoncourt, and Carl
  Dockhorn. 2019.
\newblock \href {https://doi.org/10.18653/v1/W19-4015} {{DEFT}: A corpus for
  definition extraction in free- and semi-structured text}.
\newblock In \emph{Proceedings of the 13th Linguistic Annotation Workshop},
  pages 124--131, Florence, Italy. Association for Computational Linguistics.

\bibitem[{Strubell et~al.(2018)Strubell, Verga, Andor, Weiss, and
  McCallum}]{strubell-etal-2018-linguistically}
Emma Strubell, Patrick Verga, Daniel Andor, David Weiss, and Andrew McCallum.
  2018.
\newblock \href {https://doi.org/10.18653/v1/D18-1548} {Linguistically-informed
  self-attention for semantic role labeling}.
\newblock In \emph{Proceedings of the 2018 Conference on Empirical Methods in
  Natural Language Processing}, pages 5027--5038, Brussels, Belgium.
  Association for Computational Linguistics.

\bibitem[{Vanetik et~al.(2020)Vanetik, Litvak, Shevchuk, and
  Reznik}]{vanetik-etal-2020-automated}
Natalia Vanetik, Marina Litvak, Sergey Shevchuk, and Lior Reznik. 2020.
\newblock \href {https://www.aclweb.org/anthology/2020.lrec-1.256} {Automated
  discovery of mathematical definitions in text}.
\newblock In \emph{Proceedings of The 12th Language Resources and Evaluation
  Conference}, pages 2086--2094, Marseille, France. European Language Resources
  Association.

\bibitem[{Veyseh et~al.(2020)Veyseh, Dernoncourt, Dou, and
  Nguyen}]{Veyseh2020AJM}
Amir Pouran~Ben Veyseh, Franck Dernoncourt, Dejing Dou, and Thien~Huu Nguyen.
  2020.
\newblock A joint model for definition extraction with syntactic connection and
  semantic consistency.
\newblock In \emph{Thirty-Fourth AAAI Conference on Artificial Intelligence
  (AAAI-20)}.

\bibitem[{Westerhout(2009{\natexlab{a}})}]{westerhout2009definition}
Eline Westerhout. 2009{\natexlab{a}}.
\newblock Definition extraction using linguistic and structural features.
\newblock In \emph{Proceedings of the 1st Workshop on Definition Extraction},
  pages 61--67.

\bibitem[{Westerhout(2009{\natexlab{b}})}]{westerhout-2009-extraction}
Eline Westerhout. 2009{\natexlab{b}}.
\newblock \href {https://www.aclweb.org/anthology/E09-3011} {Extraction of
  definitions using grammar-enhanced machine learning}.
\newblock In \emph{Proceedings of the Student Research Workshop at {EACL}
  2009}, pages 88--96, Athens, Greece. Association for Computational
  Linguistics.

\bibitem[{Westerhout and Monachesi(2008)}]{westerhout-monachesi-2008-creating}
Eline Westerhout and Paola Monachesi. 2008.
\newblock \href
  {http://www.lrec-conf.org/proceedings/lrec2008/pdf/783_paper.pdf} {Creating
  glossaries using pattern-based and machine learning techniques}.
\newblock In \emph{LREC 2008}.

\end{thebibliography}

\end{document}